%% file: main.tex
\documentclass[runningheads,a4paper]{llncs}
\usepackage[T1]{fontenc}

\include{config}  
\include{pythonlistings}

\usepackage{background}
\SetBgContents{Preprint}
\SetBgColor{gray}
\SetBgOpacity{0.2}
\SetBgScale{14}
\SetBgAngle{45}

\usepackage{xcolor}
\usepackage{tcolorbox}

\makeatletter
\def\watermarkoff{%
        \@sc@wm@stampfalse
}
\makeatother

\makeatletter
\def\watermarkon{%
    \@sc@wm@stamptrue
}
\makeatother

\usepackage{graphicx}
\begin{document}
\NoBgThispage
\thispagestyle{empty}

{
\Large
\noindent
PREPRINT.\\
\\
This is an author preprint version of the paper.\\
\\
See the QUATIC'21 proceedings for the final version.
\\
}

\begin{tcolorbox}[title=Cite as,colback=white,left=0pt,right=0pt]
Klikovits S., Arcaini P. (2021) KNN-Averaging for Noisy Multi-objective Optimisation. In: Paiva A.C.R., Cavalli A.R., Ventura Martins P., Pérez-Castillo R. (eds) Quality of Information and Communications Technology. QUATIC 2021. CCIS, vol 1439. Springer, Cham. \url{https://doi.org/10.1007/978-3-030-85347-1_36}
\end{tcolorbox}

\begin{tcolorbox}[title=Bibtex,colback=white,left=0pt,right=0pt]
\begin{Verbatim}[fontsize=\footnotesize]
@InProceedings{10.1007/978-3-030-85347-1_36,
  title="KNN-Averaging for Noisy Multi-objective Optimisation",
  booktitle="Quality of Information and Communications
Technology",
  year="2021",
  author="Klikovits, Stefan and Arcaini, Paolo",
  editor="Paiva, Ana C. R. and Cavalli, Ana Rosa 
and Ventura Martins, Paula and P{\'e}rez-Castillo, Ricardo",
  publisher="Springer International Publishing",
  address="Cham",
  series={CCIS},
  volume={1439},
  pages="503--518",
  isbn="978-3-030-85347-1"
}
\end{Verbatim}
\end{tcolorbox}

\clearpage

\pagebreak
\setcounter{page}{1}

%
\title{KNN-Averaging for Noisy Multi-Objective Optimisation\thanks{The authors are supported by ERATO HASUO Metamathematics for Systems Design Project (No. JPMJER1603), JST. Funding reference number: 10.13039/501100009024 ERATO. S. Klikovits is also supported by Grant-in-Aid for Research Activity Start-up 20K23334, JSPS.}
}

%
%
\author{Stefan Klikovits\orcidID{0000-0003-4212-7029} \and
Paolo Arcaini\orcidID{0000-0002-6253-4062}} %
\authorrunning{S. Klikovits, P. Arcaini} %
%
\institute{National Institute of Informatics, Tokyo, Japan \\
\email{\{klikovits, arcaini\}@nii.ac.jp}
}
\maketitle              

\begin{abstract}
\Acl{moo} is a popular approach for finding solutions to complex problems with large search spaces that reliably yields good optimisation results.
However, with the rise of \aclp{cps}, emerges a new challenge of noisy fitness functions, whose objective value for a given configuration is non-deterministic, producing varying results on each execution.
This leads to an optimisation process that is based on stochastically sampled information,
ultimately favouring solutions with fitness values that have co-incidentally high outlier noise.
In turn, the results are unfaithful due to their large discrepancies between sampled and expectable objective values.
Motivated by our work on noisy \aclp{ads}, we present the results of our ongoing research to counteract the effect of noisy fitness functions without requiring repeated executions of each solution.
Our method \emph{kNN-Avg} identifies the \acl{knn} of a solution point and uses the weighted average value as a surrogate for its actually sampled fitness.
We demonstrate the viability of \knna on common benchmark problems and show that it produces comparably good solutions whose fitness values are closer to the expected value.
\end{abstract}

\glsresetall



\section{Introduction}

In the past, \ac{moo} has proven to be a robust and reliable means in the software engineering toolbox.
Through iterative modifications of existing problem solutions, the algorithms try to approach optimal valuations. 
A \emph{fitness function} evaluates the quality of each individual solution in one generation so that the best individuals can be used to guide the next generation.
After reaching a target fitness or a given number of generations, the algorithm terminates and yields the best solutions found.

\ac{moo} algorithms such as \acp{ga} have been successfully applied to various optimisation problems, ranging from evolutionary design~\cite{Bentley1998}, biological and chemical modelling~\cite{carroll1996chemical}, to artificial intelligence~\cite{Mirjalili2019}.
\ac{moo} are particularly well-suited for modern \acp{cps} and \ac{iot} applications, given the high-dimensional search space, where combinatorial testing and exhaustive verification reach their limits.
When applied to \acp{cps}, however, a new difficulty arises. 
Many such systems suffer from sensor errors and measurement noise.
Similarly, with growing component numbers, inter-process communication causes non-deterministic behaviour due to message delays and synchronisation time variations.
As the messages' processing times differ at runtime, noisy behaviour emerges~\cite{AfzalICST20}.
Due to the noise, the reported fitness and the actually expected value (the mean over several evaluations) might vastly differ, leading to distrust of the method.

One of our lines of work is to identify problematic behaviour in \acp{ads} by creating driving scenarios through map design changes (\eg road shape), altering traffic participants (vehicles and pedestrians) and driving behaviour (\eg aggressiveness). 
However, due to noisy inter-process communication, repeated executions of the same scenario can lead to differing observations, \st the distance between two cars may vary up to several meters.
In other words, in our search for consistent collision scenarios, outliers for typical non-crash scenarios might be identified as crash scenarios.
This leads to volatile test scenarios, non-reproducible crash reports, and otherwise inconsistent scenario design.

Evidently, a naive fix is to repeatedly run each configuration and select the mean or median fitness value to check for robustness.
However, in our setting this is not possible, as a typical simulation takes 1 to 2 minutes.
Assuming 1000 total optimisation steps, re-evaluating every run five more times adds some 80 to 160 hours of simulation time to each individual scenario search, rendering this method infeasible for broad application.

In this paper, we introduce \acs{knn}-averaging to overcome the problem of lacking fitness robustness and outliers in noisy optimisation problems and provide our advances on common, theoretical benchmark problems. 
Our method \emph{kNN-Avg} uses the \ac{knn} (\selectedK being a hyper-parameter) to compute the fitness value and thereby reduce the noise.
We show that in a typical setting without noise mitigation, \acp{ga} tend to predict outlier values
and that \knna helps to overcome this problem, making the found solutions more robust.
To evaluate our approach, we adapt three well-known \ac{moo} benchmarks to noisy environments 
and quantify our results using common \acp{qi}.

\section{Multi-Objective Optimisation \& Genetic Algorithms}
\label{sec:background}

\Acl{moo} problems are a family of problems that aim to optimise multiple function values. 
Formally, the goal is to find the minimum parameter (\aka \emph{variable}) values $\vecx \in \mathbb{X}$ such that the function values of a vector of \emph{objective} functions $f$ are minimised.
$\mathbb{X} \subseteq \mathbb{R}^n$ represents the \emph{solution space}, \ie the input to $f$, and relates to the scenario parameters that we defined above. $\vecx$ is called a \emph{solution}.
\begin{equation}\label{eq:mooNoNoise}
    \min_{\vecx \in \mathbb{X}} f(\vecx) = \{f_1(x), \dots, f_m(x)\}, \forall i \in \{1, \dots, m\}, f_i: \mathbb{X} \rightarrow \mathbb{R}
\end{equation}
The output $f(\vecx) \in \mathbb{Y}$ is called the \emph{objective} value of $\vecx$; $\mathbb{Y} \subseteq \mathbb{R}^m$ being the \emph{objective space}.
An optimisation problem is called \emph{multi-objective} if $m > 1$.\footnote{In this paper, we consider \emph{unconstrained} \ac{moo} problems, meaning that $f(\vecx)$ always produces feasible output. For an introduction on constrained \ac{moo} see~\cite{fan2017comparative}.}

While for single-objective optimisation problems, a solution leading to a smaller objective value than another is considered better (\emph{dominant}),
this notion is not as clear in the \ac{moo} setting.
Here, given two solutions $\vecx_a, \vecx_b \in \mathbb{X}$, we say that $\vecx_a$ dominates $\vecx_b$ ($\vecx_a \prec \vecx_b$) iff 
$
\forall i \in \{1, \dots, n\}, f_i(\vecx_a) \leqslant f_i(\vecx_b) \wedge \exists j \in \{1, \dots, n\} \colon f_j(\vecx_a) < f_j(\vecx_b)
$.

\begin{figure}
\begin{subfigure}[t]{.43\textwidth}
\centering
    \includegraphics[width=\linewidth]{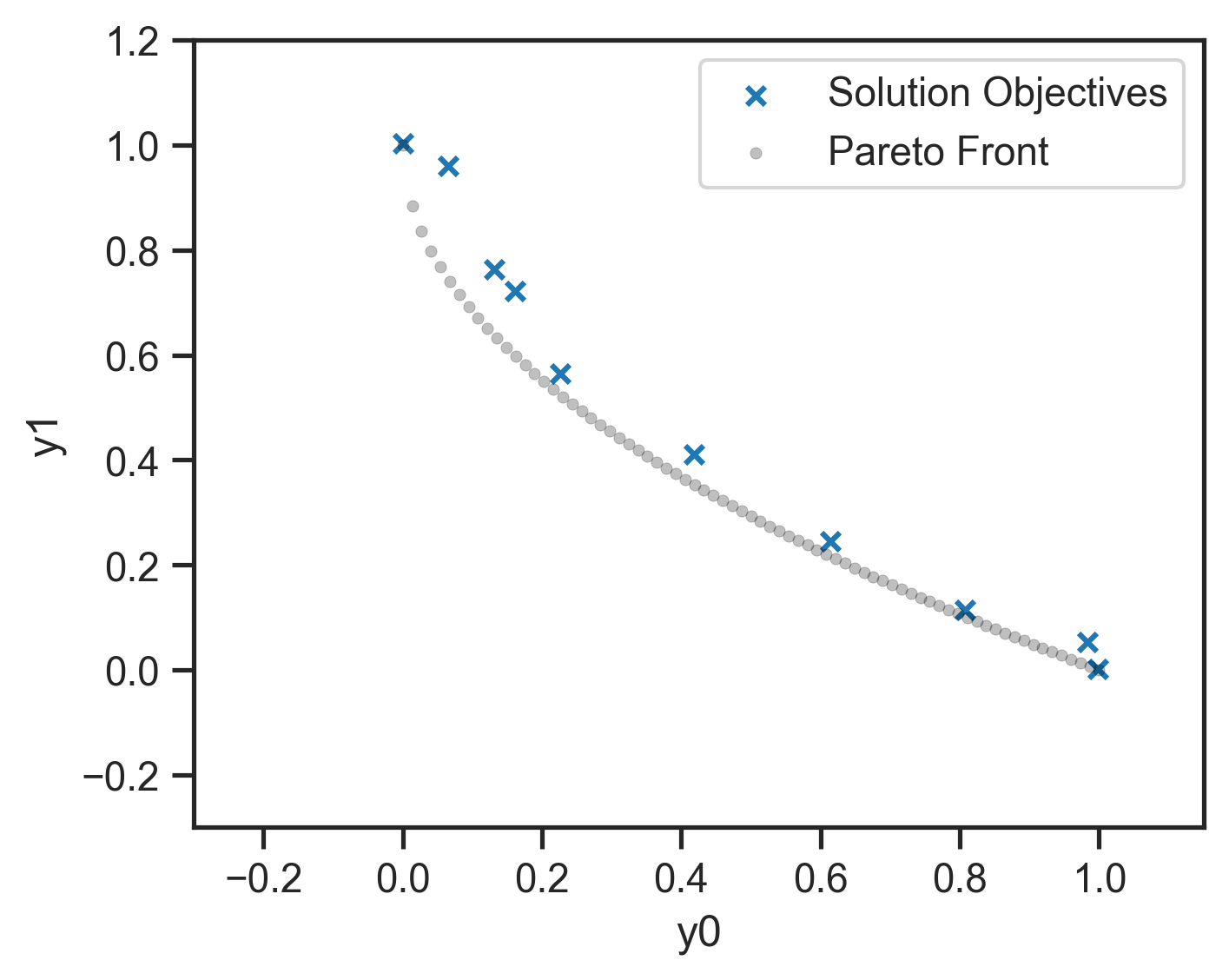}
    \caption{A solution set for the \Ac{moo} problem ZDT1 and optimal Pareto front \PF}
    \label{fig:pareto-solution}
\end{subfigure}
\hfill
\begin{subfigure}[t]{.43\textwidth}
    \includegraphics[width=\linewidth]{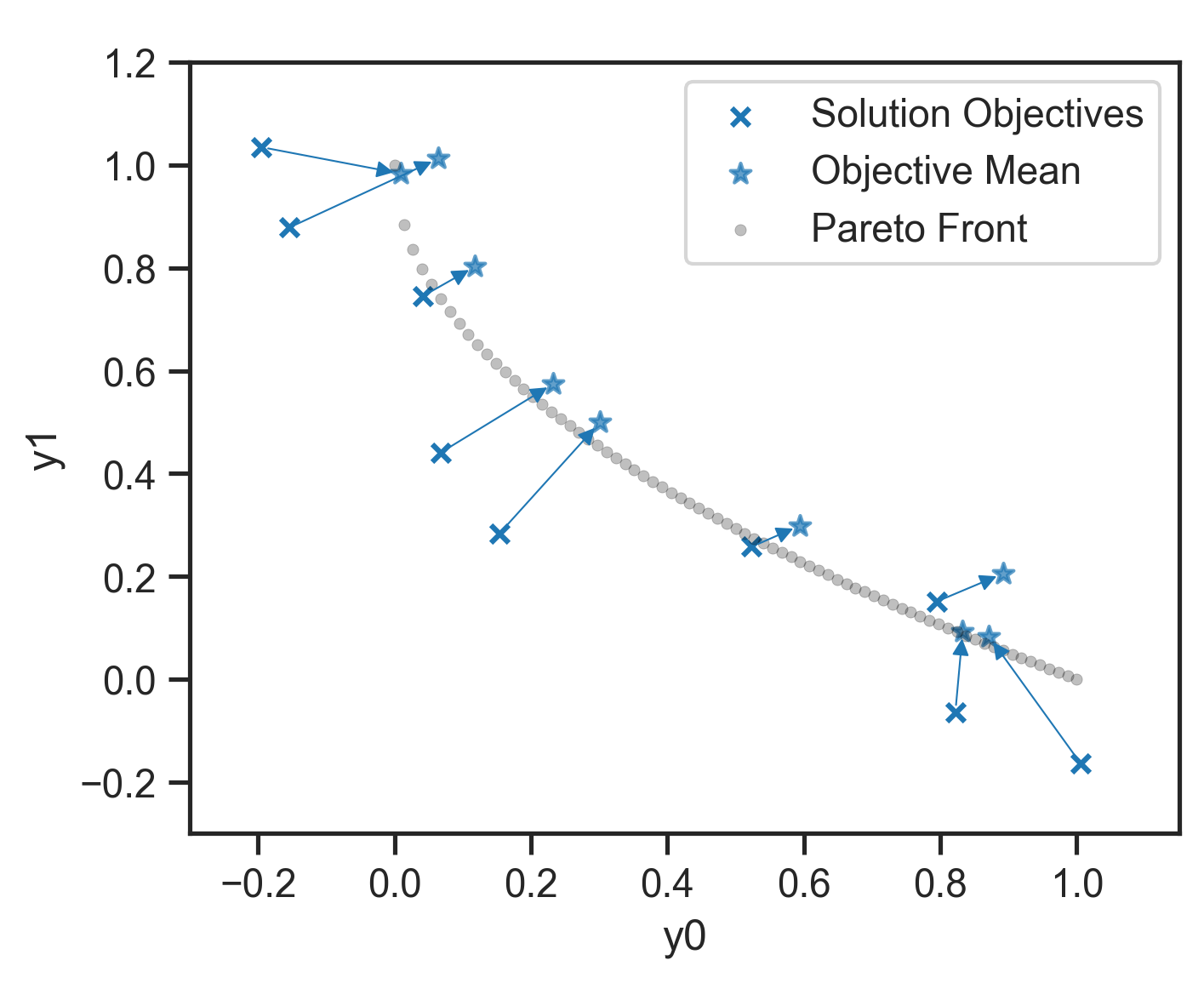}
    \caption{Noisy ZDT1$_{\sigma=0.1}$ objectives after optimisation, and the solutions' mean objectives.}
    \label{fig:noisy-solution}
\end{subfigure}
    \caption{Left: search problem without noise; Right: Noisy search problem}
    \label{fig:figure1}
\end{figure}

A solution is \emph{Pareto optimal} iff there is no other solution that dominates it.
A Pareto set  $\PS \in \mathbb{X}$ is the set of all Pareto optimal solutions, a Pareto-front PF$^{*}$ is the image of the Pareto set in the objective space.
Roughly speaking, \PS is the set of solutions where we cannot optimise one $f_i$-dimension without having to reduce optimality in another one.
This means, however, that there might be numerous Pareto optimal solutions for some problems. 
\cref{fig:pareto-solution} shows a solution set for the benchmark problem ZDT1~\cite{ZDT2000} alongside its Pareto front \PF.
Note that each value of the solution set is Pareto optimal.

\vspace{5pt}

\noindent{\bf Quality Indicators (\acs{qi})}
\Acp{qi} allow an estimation of a \ac{moo}'s performance.
Numerous \acp{qi} have been proposed~\cite{li2019quality}.
In this paper, we focus on two of the most common ones.
\begin{compactdesc}
\item \emph{\Ac{hv}} It calculates the hypervolume between a given reference point in the objective space and each of the solutions' objective values. 
Typically, it uses a reference that is worse than all expected objective values. 
Thus, the \ac{hv} increases as the solutions approach the Pareto-optimality.


\item \emph{\Ac{igd}} \ac{igd} calculates the average Euclidean distance from each \PF-point to its closest solution.
\ac{igd} decreases when approaching \PF.
\end{compactdesc}

\vspace{5pt}

\noindent{\bf \Acfp{ga}}
\Acp{ga}~\cite{Eiben2015} are a subfamily of \acp{ea} whose working principle is based on the iterative creation of \emph{populations} (sets of solutions, \aka \emph{generations}).
An initial population can be created from random samples, and all following generations are based on the best candidates from their predecessors.
This is achieved by ranking the solutions according to their fitness values $f(\vecx)$ and applying \emph{operators} to the best ones:
\emph{selection} keeps the good ones,
\emph{crossover} produces new solutions by ``mating'' two others, and \emph{mutation} creates a slightly altered version of an existing one.

\section{Noisy \ac{moo} and K-Nearest-Neighbour Averaging}
\label{sec:methodology}


We use the term \emph{noisy \ac{moo}} to refer to problems whose fitness function $f(x)$ is non-deterministic.
The formal problem statement, introduced in Eq.~\ref{eq:mooNoNoise}, can be updated to 
\begin{equation}
  \min_{\vecx \in X} f(\vecx) = \{f_1(\vecx) + \delta_1, \dots, f_m(\vecx) + \delta_m\}
\end{equation} 
where each $\delta_i$ is a noise value sampled from a distribution.
In many real-world systems, as well as common benchmark studies, the noise in the system is Gaussian-distributed~\cite{goh2009evolutionary}.
To make our results comparable, we follow this lead, and therefore, for the rest of this paper, we fix that $\delta_i$ is 
sampled from a normal distribution with mean $\mu = 0$ and a standard deviation $\sigma$.
We use $\sigma$ as a variable and alter it depending on the particular benchmark.
For noisy systems, we indicate the problem's noise $\sigma$-parameter as subscript, \eg ZDT1$_{\sigma = 0.1}$, as in \cref{fig:noisy-solution}.

\vspace{3pt}

\noindent\paragraph{\bf Noise Effect}
When introducing noise into an optimisation problem, the effect is that \PF no longer represents 
the optimal objective values but instead marks the mean objective values for \PS.
Given a solution $\finalSol \in \PS$ and high enough repetitions, the mean value should be placed near the solution's expected (non-noisy) objective.

Interestingly though, when executing a \ac{ga} on such a problem, the typical result (see also ~\cite{goh2007investigation}) is 
that the algorithm computes a set of solutions that approaches \PS, but predicts objective values with high negative noise, 
falsely indicating extraordinarily good fitness values.
Thus, in the final solution set \finalPF, the \ac{ga} ``pushes'' the reported fitness value far beyond \PF, as can be seen in \cref{fig:noisy-solution}.
The result is that the objective values do not match the actual distributions of the solutions' expected mean objective values, 
and in fact over-estimate the quality of the \finalPF.
\cref{fig:noisy-solution} displays both the sampled solutions' objective values as well as these solutions' mean objective values.
\finalPF was obtained by running 100 iterations of the \ac{ga}-optimisation of the noisy ZDT1$_{\sigma = 0.1}$ problem.
In the figure, we can observe several properties:
\begin{compactenum}
    \item The objective values of the found non-dominated solution set \finalPF (marked by crosses) actually significantly surpass \PF.
    \item By evaluating $\finalSol \in \finalPF$ repeatedly (or by removing the noise), we reveal the actual mean objective values, \ie the centre points of the noisy objective distributions (marked by asterisks).
    \item Some of the mean objective values are clearly dominated by others. Thus, evaluating $\finalSol \in \finalPF$ will likely lead to non-dominated solutions.
    \item The observed objective values are typically based on large noise values, \ie they are outliers.
    \item On average, the distance between observed objectives and the mean objectives (indicated by the arrows) is rather large ---  even exceeding $\sigma$. Thus, the obtained solutions' fitness is misleading.
    \item Even though the solutions' objective values do not match the mean objective values, the solutions' mean objective values approach \PF. This means that while we cannot expect fitness values to be as good as predicted by the optimisation process, on average the solution values will still be good.
    \item It appears that the noise causes the search to stop further improvement, since outlier noise suggests the greatest possible fitness was already reached.
\end{compactenum}
Ideally, the solutions' mean objectives would align with \PF, \st this result would produce the best fitness on average.

\vspace{3pt}

\noindent\paragraph{\bf Research Goal}
Indeed, in \cref{fig:figure1} where we know that the noise (induced by $\sigma$) is constant throughout the search space and as we know \PF, we can easily see the offset.
However, for unknown problems, problems with potentially fluctuating noise magnitudes or similar situations, we should aim to produce solutions whose sampled objective values are close to the mean objective values. 
This way, we know that the individuals' objectives are reliable and the obtained values representative.
In other words, our goal is to push the crosses in \cref{fig:noisy-solution} closer towards the asterisks, while keeping the asterisks close to the ideal Pareto front \PF.

\noindent\paragraph{\bf Mean Offset \meanErr} We introduce a new measure \meanErr to calculate the mean Euclidean distance between \finalPF's reported objective values $f(\finalSol), \finalSol \in \finalPF$ --- as obtained by the search ---  and the mean expected objective values $\tilde{f}(\finalSol)$ --- obtained \eg by taking the mean of repeated invocations of $f$ on $\finalSol$. 
\meanErr thus measures the average length of the arrows in \cref{fig:noisy-solution}.
The rest of this section describe our method which leverages the weighted average of the \acl{knn} to robustify the sampling process and reduce \meanErr.
\begin{equation}\label{eq:meanOffset}
\meanErr = \frac{
    \sum_{\finalSol \in \finalPF}\sqrt{(f(\finalSol) - \tilde{f}(\finalSol))^2}
}{\mid \finalPF \mid}
\end{equation}

\subsection{KNN-Averaging}\label{sec:knnAveraging}

Theoretically, inputs to noisy fitness functions can be repeatedly evaluated and the sampled values averaged.
Given high enough repetition frequency, this mean objective value should approach the actual expectation value.
The problem of this method is the cost involved in the re-sampling of values, which can be significant for complex systems such as automated driving simulators, even rendering the search effectively infeasible.

Our novel approach \emph{\knna} aims to approximate the re-sampling process and decrease the objective values noise, without actually performing the costly repeated sampling.
Our concept is based on the hypothesis that solutions that are close to one another in the solution space will produce similarly distributed objective values.
Thus, the goal is to identify previously evaluated solutions close to the currently sampled point and use their fitness values to help decrease the impact of outliers.

\paragraph{Standardised Euclidean Distance}

\Acf{ed} is one of the go-to distance measures when it comes to evaluating the proximity of multi-dimensional data points, providing the ``shortest direct distance'' between two points.
On closer inspection, however, we notice that often the values depend on the individual dimensions' units.
Thus, the Euclidean distance of values expressed in kilometres is vastly different than the one in metres.
Moreover, the measure is not robust against large magnitude differences. When one dimension expresses vehicle size in metres, but the other travelled distances in kilometres, the second dimension will probably dominate the measure.

Our goal is thus to use a measure that normalises over the actually used search space to avoid such tilting of the distance measure.
We therefore use the \acf{sed}, which normalises the values by relating the value difference to the dimension's variance within the total set of values in that dimension.

\begin{definition}[\Acl{sed}]
The \acf{sed} of two vectors is the square-root of the sum of the squared element-wise difference divided by the dimension's variance $\sigma_{i}^2$:
\begin{equation}
\mathit{sed}(\vecx_1,\vecx_2)= \sqrt{\sum_{i=1}^N  {\frac{(x_{1,i} - x_{2,i})^2}{\sigma_{i}^2}}},
\end{equation}
\end{definition}
Note that the difference to the ``common'' \acl{ed}, \ie the division by the dimension's variance $\sigma_{i}^2$, places each dimension in relation to the global value spread.
This robustifies \ac{sed} against magnitude changes in the solution space and we can---in theory---even change, \eg the units from \texttt{km} to \texttt{m}.

\subsection{Hyper-Parameters of \knna}
Next to the choice of distance measure, the \knna algorithm offers several other configuration parameters. 
Here, we will briefly outline some of the questions that have to be answered before executing the algorithm.

\begin{compactdesc}
\item[\textbf{How big should \selectedK be?}]
    Intuitively, we might want to include as many neighbours as possible to minimise the noise factor. 
    The problem is, however, that higher \selectedK-values take more distant neighbours into account, which are less similar to the solution under investigation.
    Typically, the answer also depends on the shape of the search space and the standard deviation of the noise.
\item[\textbf{When does a neighbour stop being ``near''?}]
    Especially at the beginning of the search when the sampling history is still sparse, a \ac{knn} algorithm without a \emph{cutoff} limit might select rather distant neighbours for the averaging. To avoid this, we use a \emph{maximum distance} \maxDist to limit the \ac{knn} search to the local neighbourhood.
\item[Should close neighbours weigh more?] 
    By default, our hypothesis is that close neighbours produce more similar value distributions than those further away.
    Based on an informal initial exploration, our algorithm uses the squared \ac{sed} to decrease a neighbours impact weight.
    Nonetheless, this raises the question as to how those weights are shaped in general.
    We leave this more general question as future works.
\end{compactdesc}

\vspace{5pt}

We performed an upfront literature search, but it did not reveal conclusive answers or best practices for good choices on these hyper-parameter values. 
Thus, \selectedK and maximum distance \maxDist are left variable.
The experiments section explores the relationship between these hyper-parameter values and the performance of the \ac{moo} process.
The weight measure on the other hand was fixed as the inverse-square of the \ac{sed}.
Before choosing squared, we also experimented with \emph{linear} and \emph{uniform} distance weights, but did not observe as good results.

\subsection{Algorithm}
Based on the \knna approach and the identified hyper-parameters we implemented the algorithm,
as displayed in \Cref{lst:knn-algorithm}.
\begin{figure*}[!tb]
\begin{lstlisting}[language=iPython, mathescape=true, % float, % float makes it behave like a figure...
caption={KNN-Evaluation algorithm in (Python-ish) pseudo-code},label={lst:knn-algorithm}]
"""Store the solutions with actually sampled values."""
HISTORY = list()  # global store of solutions <@\label{line:history}@>
"""Calculate KNN-averaged objectives for a list of solutions."""
def knn_evaluate(population: List[Solution], KNN: int, MAX_DIST: float)  -> List[Solution]:
  # standard evaluation of each solution; add results to the history
  sampled = [evaluate(solution) for solution in population]  <@ \label{line:sampling} @>
  HISTORY.extend(sampled) <@\label{line:history-extension}@>
  # calculates the list of variance (one per variable dimension)
  variances = get_variances(HISTORY) <@ \label{line:variances} @>  
  # calculate solutions with KNN-averaged objectives
  knn_solutions = list()
  for solution in population:  <@\label{line:iteration}@>
    # store a map of solution-to-SED 
    distances={other:SED(solution, other, variances) for other in HISTORY}<@ \label{line:distance-calc} @>
    # remove those that are larger than MAX_DISTANCE
    knn = {sol: val for sol, val in knn.items() if val <= MAX_DIST) <@\label{line:max-dist-cut}@>
    # sort by distance and limit to KNN values
    knn = distances.sort(by=distances.values())[:KNN]  <@\label{line:knn-cut}@>
    # calculate the weights as the square of distance values; 
    # negate and add MAX_DIST so sol itself is largest
    weights = math.square(knn.values()) * -1 + MAX_DIST <@\label{line:weight-calc}@>
    objs = [sol.objs for sol in knn] <@\label{line:mean-calc1}@>
    avgs = weighted_mean(objs, weights, column=True)  <@\label{line:mean-calc2}@>
    # append to the list of returned solutions
    knn_solutions.append(Solution(solution.vars, avgs))
  return knn_solutions

"""Data-class for solutions and corresponding objective values."""
class Solution(object):  <@\label{line:solution}@>
  vars = list()  # variables will be filled by the GA
  objs = list()  # objectives are filled by us
\end{lstlisting}
\end{figure*}
The algorithm is presented in a Python-like pseudo-language and makes references to 
``mock'' functions such as \py{get_variances}. 
The actual implementation uses common Python libraries such as \py{numpy} and \py{pandas}. 

The listing's \py{knn_evaluate}-function is meant to replace the native \py{evaluate}-function of a \ac{ga} implementation. 
It is called once per generation and takes as input a list of solutions, as well as the \py{KNN} and \py{MAX_DIST} hyper-parameters.
The output is a list of \py{Solution} objects, each containing variables and \ac{knn}-averaged objective values.
Specifically, each population cycles through the following steps:
\begin{compactenum}
\item The algorithm first invokes the (noisy) default \py{evaluate} method (\Cref{line:sampling}). This can be \eg the triggering of a simulator.
\item After this evaluation, the solutions (now containing the sampling results) are added to the global \py{HISTORY} (\Cref{line:history} and \Cref{line:history-extension}).
\item In the next step, the \py{variances} are calculated for each solution dimension separately (\Cref{line:variances}). The variances are later used to calculate the \ac{sed}.
\item Then, each solution iterates through the following steps:
\begin{compactenum}
    \item Calculate the solution's \acp{sed} to each other solution in the global history --- including itself and all other newly generated solutions (\Cref{line:distance-calc}).
    \item Select all solutions closer than \py{MAX_DIST} (\Cref{line:max-dist-cut}).
    \item Sort solutions by the \ac{sed} and only keep the \acl{knn} (\Cref{line:knn-cut}).
    \item Calculate the \py{weights} of each \py{knn} as the negative square of its \ac{sed} (\Cref{line:weight-calc}). Then add \py{MAX_DIST} to make all values positive.
    \item Calculate the objective values as the weighted mean values. Note that since \py{objectives} is a two-dimensional list, we have to specify \py{column}-wise aggregation, such that each dimension is averaged separately (\Cref{line:mean-calc1} -- \ref{line:mean-calc2}). 
    The solution variables and averaged objectives are used to append a new \py{Solution} to the list that will be returned to the algorithm.
\end{compactenum} 
\end{compactenum}

\begin{figure}
    \centering
    \includegraphics[width=.75\linewidth]{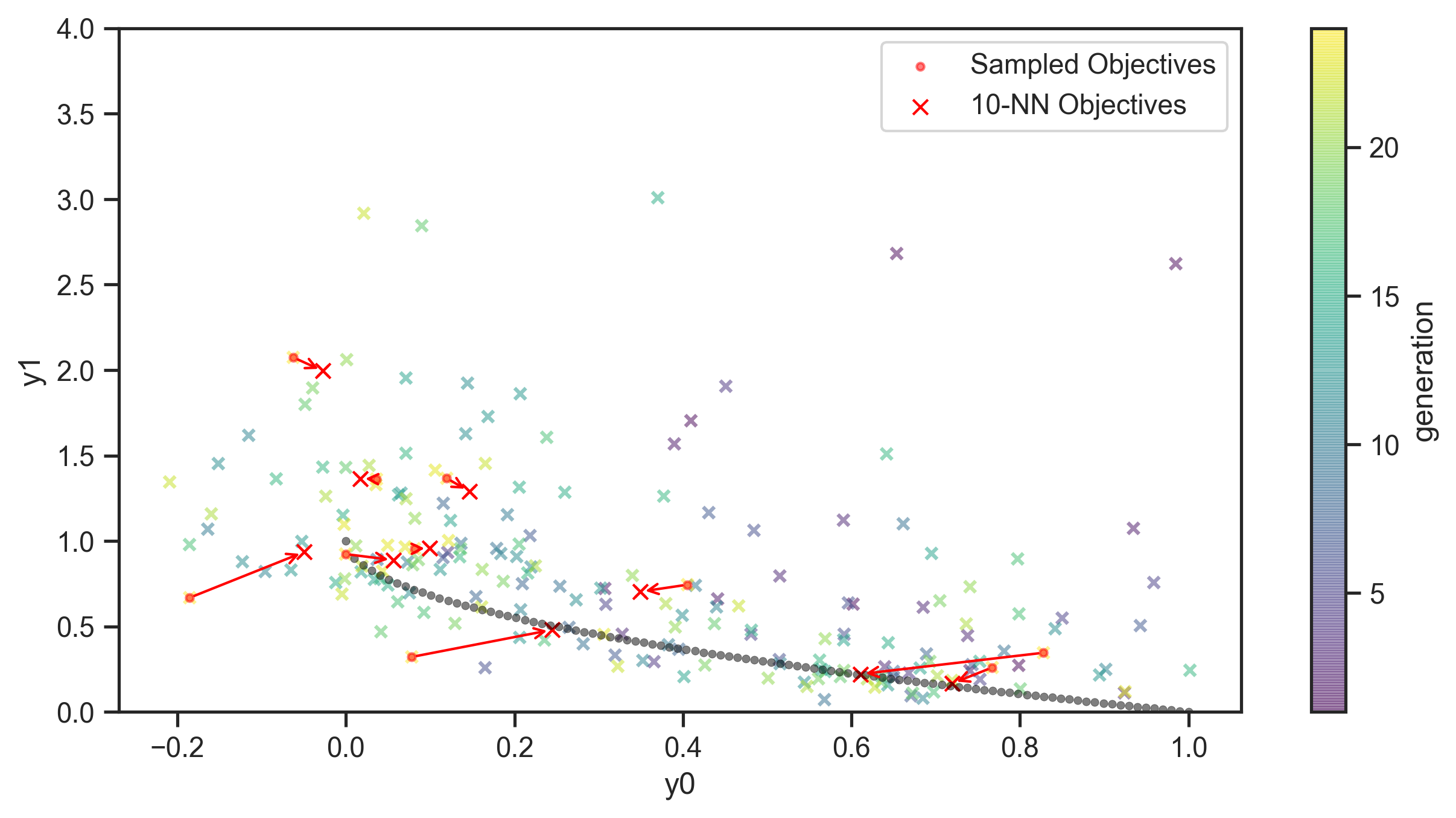}
    \caption{10-NN Averaging at Generation \#25 for ZDT1$_{\sigma=0.1}$. Arrows connect sampled and respective 10-NN-averaged objective values.}
    \label{fig:knn-arrows}
\end{figure}

\paragraph{\knna Effect} 
The effect of the algorithm is visualised in \cref{fig:knn-arrows}. 
It shows the \knna approach at the evaluation of the 25th generation of a search problem.
The evaluation history is shown as coloured crosses. 
Red dots display the sampled objective values, arrows indicate the corresponding objective values after \knna (red crosses).

\section{Evaluation}
\label{sec:evaluation}

This section describes the experiments and analyses we performed to answer the following research questions (RQs):
%
\begin{compactitem}
\item {\bf RQ1}: Can \knna mitigate the outlier-effect of noisy \ac{moo} and bring the reported and mean objective values closer together (\ie reduce \meanErr)?
\item {\bf RQ2}: How do the solutions of \knna compare to those of the baseline in terms of optimality (measured by \acfp{qi})?
\item {\bf RQ3}: Does the level of noise have an impact on whether \knna is able to produce good results?
\item {\bf RQ4}: What influence does the choice of hyper-parameters have on the efficiency of the approach?
\end{compactitem}

\subsubsection{Experimental Setup}

To evaluate the \knna method, we implemented a set of noisy benchmarks.
The implementation is based on the \pymoo Python library~\cite{pymoo}, 
which provides a flexible framework for evaluation of \ac{moo} problems and algorithms.

For the \knna evaluation, we developed a \py{KNNAvgMixin}-wrapper that we can dynamically add to existing \pymoo problems. 
The wrapper serves two purposes: 
\begin{inparaitem}[]
\item First, it modifies the wrapped problem
and artificially injects noise into the evaluated solutions.
\item Second, it implements the \knna algorithm to counteract the effects of the noise.
\end{inparaitem}%
The wrapper also stores the full evaluation history and adds data logging for our analysis. 
It can be flexibly added to existing benchmark problems, \eg those already available in \pymoo.

Using this class, we ran experiments on three benchmark problems: ZDT1, ZDT2 and ZDT3\cite{ZDT2000}. 
These artificial \ac{moo} benchmarks aim to minimise two-function objectives, based on a variable number of up to 30 inputs.
In total, we selected six individual settings with multiple values each, to avoid biasing our algorithm and also obtain an overview of the influence of various hyper-parameters (RQ4).
Following guidelines~\cite{Arcuri2011}, we executed 30 repetitions in each setting to avoid statistical fluctuations and gain enough data for our later analysis.
\cref{tbl:experiment-settings} displays the experiment settings and the values that each one may take.
The total number of experimental settings is given by the Cartesian product of the settings' values. In total, we executed $67{,}500$ individual optimisation runs.

As \ac{moo} method we use NSGA-II, configured with a random initial population, simulated binary crossover (probability 0.9) and polynomial mutation (probability 1.0). The search was run for 100 generations using population size 10 or 20 (see \cref{tbl:experiment-settings}).
\begin{table}[!t]
    \centering
    \setlength{\tabcolsep}{1em}
    \caption{Configurations of benchmarks and search algorithms}
    \label{tab:expSettings}
    { \scriptsize
    \begin{tabular}{ccc}
    \toprule
        Label & Values & Comment \\ 
        \midrule
        Problem & ZDT1, ZDT2, ZDT3 & \textit{Benchmark name} \\
        Num Variables & 2,\quad 4,\quad 10 & \textit{Benchmark configuration} \\
        $\sigma$ (Noise Std. Dev.) & 0.00, 0.05, 0.10, 0.25, 0.50 & \textit{Benchmark Noise} \\[.5ex] \hline
        Population Size & 10,\qquad 20 & \textit{\ac{ga} setting} \\
        \selectedK (Num Neighbours) & 10, 25, 50, 100, 1000 & \textit{\knna hyper-parameter} \\
        \maxDist maximum distance & 0.25, 0.5, 1.0, 2.0, 4.0 & \textit{\knna hyper-parameter}  \\
        \bottomrule
    \end{tabular}
    }
    \label{tbl:experiment-settings}
\end{table}


\subsection{Experimental Results}

In order to evaluate the effectiveness of the approach, we proceeded as follows.

For each experiment, we took the set of optimised solutions \finalPF as computed by the \ac{ga}; then for each solution $\finalSol \in \finalPF$, we computed its ``mean'' objective value \fixedSol; all solutions \fixedSol constitute the ``adjusted'' solution set \fixedPF.
Based thereon, we computed the following metrics:
\begin{inparaenum}[(i)]
\item \meanErr, the average error in estimation between the objective values of solutions in \finalPF and the adjusted ones in \fixedPF (see \cref{eq:meanOffset});
\item \meanQI{\HV} and \meanQI{\IGD}, the \acl{hv} and \ac{igd} (see \cref{sec:background}) of the mean objectives $\fixedSol \in \fixedPF$.
\end{inparaenum}

\begin{figure}[!tb]
\centering
\includegraphics[width=.9\linewidth]{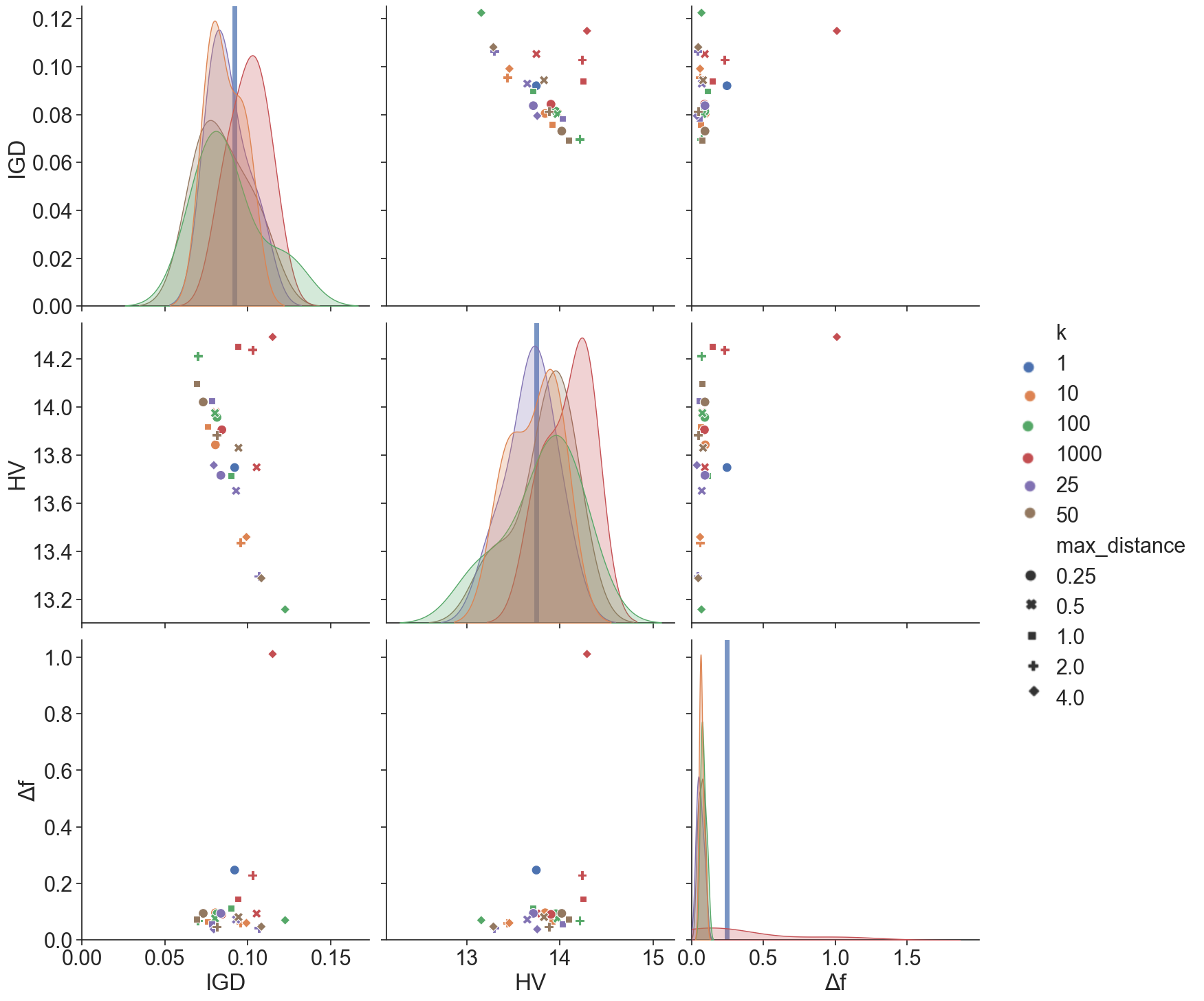}
\caption{Pairplot of average \ac{qi} and \meanErr of 30 runs for ZDT1$_{\sigma=0.1}$. Diagonal plots show kernel density estimates. Baseline \baseline ($k = 1$; no averaging) is shown as a vertical line.}
\label{fig:pairplot}
\end{figure}

As an example, \cref{fig:pairplot} shows \meanQI{\HV}, \meanQI{\IGD}, and \meanErr for the ZDT1$_{\sigma=0.1}$ experiment with 2 variables, grouped by \selectedK and maximum distance \maxDist, \ie by the hyper-parameters we want to evaluate\footnote{The implementation of our algorithm and the plots of all other experimental settings are available online  \url{https://github.com/ERATOMMSD/QUATIC2021-KNN-Averaging}.)\label{foot:repo}}.
The plots show the mean results of 30 individual NSGA-II optimisation runs after 100 generations with population size 10.
The individual plots show two metrics plotted against each other, while the diagonal plots display the kernel density estimate\footnote{\url{https://www.mvstat.net/tduong/research/seminars/seminar-2001-05}} by \selectedK value. 
Roughly speaking, this is a distribution of where the individual metrics (grouped by \selectedK) are along the x-axis. 
Thus, for \meanQI{\IGD} and \meanErr, lower (left) is better, while for \meanQI{\HV} higher (right) is advantageous. Further, a flat, wide density shows a spread of values along the axis, while a short, high curve signifies a concentrated values.
The baseline represents the standard (non-averaging) noisy \ac{moo}, where only the sampled value is chosen without any neighbours(\ie $\selectedK=1$). 
It is is drawn as vertical blue line. Note, that a comparison to other solution approaches to noisy \ac{moo} is left as future work.

We see from the positions of the data points, as well as from the density plots, that for most \selectedK values (except $k$$=$$1000$) the \meanQI{\IGD} is lower than \baseline (blue line). For \meanQI{\HV} the value is higher, except $k$$=$$25$. Of particular interest is \meanErr, which shows a significant discrepancy between the \knna evaluations and \baseline, in favour of \knna.

Thus, we conclude that for this particular benchmark, the \knna approach produces better results on average than \baseline.
We report in the online repository\textsuperscript{\ref{foot:repo}} the detailed analyses of all the 45 benchmarks. In the following, we perform an overall analysis to assess how the proposed approach performs in general.

We calculated the experiments results' average (grouped by \selectedK and \maxDist) and compared them with the results of the baseline \baseline, across all experiment configurations (all combinations of \emph{Problem}, \emph{Num Variables}, $\sigma$, and \emph{Population size}).
By following guidelines for conducting experiments with randomised algorithms~\cite{Arcuri2011}, we compared them using the Wilcoxon signed rank test for the statistical significance (at significance level $\alpha$$=$$0.05$), and the Vargha-Delaney's \Atwelve as effect size.
\cref{table:expResults} reports the results of the statistical tests, displaying whether \knna's performance is statistically significantly better, equal, or worse than the \baseline's.
\input{tables/tableStatTestsFiltered}
\cref{tab:resultsNoise0} reports the results for the benchmarks with no noise, \cref{tab:resultsNoise0.05}-\ref{tab:resultsNoise0.5} the results for the benchmarks having a specific noise level \noise{a}, and \cref{tab:resultsAllNoise} the results by benchmarks with any level of noise.

\subsection{Evaluation}

\noindent{\bf RQ1: Can \knna mitigate the outlier-effect of noisy \ac{moo} and bring the reported and mean objective values closer together?}
From \cref{tab:resultsNoise0}, it is clear there is no advantage in using \knna for non-noisy systems. Indeed, the approach computes approximated fitness values, although the sampled ones are already precise.
Instead, for any noise level (\Cref{tab:resultsNoise0.05}-\ref{tab:resultsNoise0.5}) and most experimental settings, statistically, \knna outperforms \baseline in terms of \meanErr. This clearly shows that \knna succeeds in producing solutions whose sampled objective values are closer to their mean values.


\vspace{10pt}

\noindent{\bf RQ2: How do the solutions of \knna compare to those of \baseline in terms of optimality (\ie in terms of quality indicators)?}
For non-noisy benchmarks (\cref{tab:resultsNoise0}), \knna is worse than \baseline in term of solution quality; this effect is expected, as the approach perturbates the fitness value when it is not needed. Nonetheless, for any noise level (\Cref{tab:resultsNoise0.05}-\ref{tab:resultsNoise0.5}), we observe that there are several hyper-parameter configurations where \knna produces equal or better results for both \meanQI{\HV} and \meanQI{\IGD}.


\vspace{10pt}

\noindent{\bf RQ3: Does the level of noise have an impact on whether \knna is able to produce good results?}
With increasing noise level, \knna's results improve. For noise \noise{0.5} (\cref{tab:resultsNoise0.5}), none of the settings of \knna produces worse solutions. This shows that our approach is particularly efficient for highly noisy systems.

\vspace{10pt}

\noindent{\bf RQ4: Which is the influence of the method hyper-parameters on the efficiency of the approach?}
\knna is initialised with two parameters, the number of neighbours \selectedK, and the maximum distance \maxDist. The results suggest that there is no big influence of the used \selectedK. For \maxDist on the other hand, lower values usually provide better results across noise levels. This is reasonable, as smaller values of \maxDist make the approach more conservative and avoid averaging too different values.

\subsection{Threats to Validity}\label{sec:threatToValidity}

The validity of \knna could be affected by some threats. We discuss them in terms of \emph{construct}, \emph{conclusion}, \emph{internal}, and \emph{external validity}.

\noindent{\bf Construct validity}
One threat is that the evaluation metrics may not reflect the object of the investigation, that is, the ability of \knna to produce solutions with low \meanErr and still have a good quality in terms of the objective functions. Furthermore, it may be that the newly introduced measure \meanErr is not a faithful measure for robustness in this context.
As different \acp{qi} may give different results (in terms of solution ranking), we used two distinct ones~\cite{li2019quality} to avoid biasing; of course, many more indicators could have been used.
We further applied this result on several benchmarks to judge the results.

\noindent{\bf Conclusion validity}
Different factors can affect the ability to draw definitive conclusions; one of these is the random behaviour of the search algorithms. To mitigate such a threat, we executed each experiment 30 times, as suggested in a guideline on conducting experiments with randomised algorithms~\cite{Arcuri2011}. Still following~\cite{Arcuri2011}, we compared the results of different versions of \knna and of the baseline approach by using suitable tests that account both for statistical significance and effect size.

\noindent{\bf Internal validity}
One threat could be to wrongly identify a causal relationship between the usage of \knna and the improvement in \meanErr. To mitigate, we carefully tested the implementation, and we make it available for inspection and experiments reproduction.

\noindent{\bf External Validity}
\knna has been experimented on 45 benchmark models, varying in objective functions, variable numbers, and noise. The benchmarks are commonly used in the \ac{moo} community to assess search algorithms. However, more experiments are needed to claim the generability of the approach, possibly using more complex \acp{cps} affected by noise, such as \acp{ads}. This is left as future work.

\section{Related Works}
\label{sec:relatedworks}



Other approaches have been proposed for handling noise in multi-objective optimisation; see~\cite{goh2009evolutionary} for a survey.
Early works~\cite{Fitzpatrick1988} suggest performing multiple evaluations of the fitness functions and try to understand the sufficient number of evaluations; such approaches may not be applicable in practice when evaluating the fitness functions is expensive (\eg using \ac{ads} simulators).

Park and Ryu~\cite{ParkGECCO2011} propose to handle the noise by performing multiple evaluations of the solutions over several different generations. The approach differs from ours, as we rely on the average fitness values of the \ac{knn}, while they calculate the average of multiple re-executions of the same solution.

Other methods~\cite{Hughes2001,Teich2001}, instead, propose to modify the ranking method to take the system noise into account. The main problem of these works is that they make prior assumptions on the distribution of objective function values; \knna, instead, makes no assumption on the noise distribution and tries to discover it at runtime.

The closest approach to ours is proposed by Branke~\cite{Branke1998}. It considers averaging as one of the 10 possible ways to estimate fitness. However, the approach differs in several ways:
\begin{inparaenum}[(i)]
\item the approach only applies to single-objective problems;
\item the distance function does not consider the different dimensions (as we do with SED, see \cref{sec:knnAveraging});
\item it takes the whole population into account, instead of limiting the averaging to the \ac{knn}.
\end{inparaenum}

\section{Conclusion \& Future Works}
\label{sec:conclusion}

\glsresetall

This paper presents a novel approach to \ac{moo} of noisy fitness functions. In such settings, \ac{moo} methods such as \aclp{ga} tend to wrongly rely on outlier values to guide the optimisation. 
This results in the problem that the fitness of the reported solutions differs significantly from the expected fitness values obtained by re-running the solutions, which harms trustworthiness.
A naive fix would be to repeatedly sample a solution and calculate the mean of observed values, which can easily become very costly for more complex systems.

We present an approach for reducing the noise while avoiding re-sampling. 
Our \emph{kNN-Avg} algorithm works by keeping the history of all evaluated solutions and calculates the weighted mean of the \ac{knn}.
We show the details of our implementation and provide an experimental evaluation based on three common benchmark problems, modified with different noise levels.
The results indicate that our \knna method indeed succeeds in reducing the discrepancy between the solution's fitness and the actual target fitness, thereby increasing the trustworthiness of results.

In future, we plan to extend our approach in several directions.
First, we plan on applying \knna to more benchmark problems, including constrained ones, and evaluate its performance on different types of noise.
Next, we are in the process of evaluating the method in a scenario generation setting for an \acl{ads}.
Finally, we want to investigate the algorithm's hyper-parameters and test if any correlations between problem, noise level and algorithm configuration exist.

\bibliographystyle{splncs04}
\bibliography{bibliography}
%
\end{document}

%% file: config.tex
\usepackage[svgnames,table]{xcolor}
\definecolor{darkblue}{rgb}{0, 0, 0.7}

\usepackage{fancyvrb}

\usepackage{amsfonts}
\usepackage{amssymb}
\usepackage{amsmath}
\usepackage{times}

\usepackage{caption}
\usepackage{subcaption}
\usepackage{pifont}
\usepackage{booktabs}

\usepackage{wrapfig}

\usepackage{paralist}

\usepackage{listings}
\usepackage[colorlinks=true,
pagebackref=false,
urlcolor=blue!50!blue,
citecolor=blue!50!blue,
linkcolor=darkblue]{hyperref}

\usepackage[capitalise,nameinlink]{cleveref}
\crefname{lstlisting}{listing}{listings}
\Crefname{lstlisting}{Listing}{Listings}
\crefname{line}{Line}{Lines}
\Crefname{line}{L}{L}
\crefname{section}{§}{§§}
\Crefname{section}{§}{§§}

\usepackage{xspace} 

\newcommand{\aka}{a.k.a.\xspace}
\newcommand{\eg}{e.g.\xspace}

\newcommand{\ie}{i.e.\xspace}
\newcommand{\st}{s.t.\xspace}

\newcommand{\wrt}{w.r.t.\xspace}

\newcommand{\vecx}{\ensuremath{\vec{x}\xspace}}

\newcommand{\knna}{kNN-Avg\xspace}
\newcommand{\PS}{PS\ensuremath{^{*}}\xspace}
\newcommand{\PF}{PF\ensuremath{^{*}}\xspace}

\newcommand{\pymoo}{\texttt{pymoo}\xspace}

\newcommand{\knnAvg}[2]{\ensuremath{\mathtt{KNN}_{#1}^{\ensuremath{#2}}}\xspace}
\newcommand{\baseline}{B\xspace}
\newcommand{\noise}[1]{\ensuremath{\sigma={#1}}\xspace}
\newcommand{\HV}{HV\xspace}
\newcommand{\IGD}{IGD\xspace}

\definecolor{badred}{RGB}{240,0, 0}
\definecolor{goodgreen}{RGB}{0, 216, 0}

\newcommand{\meanQI}[1]{\ensuremath{\widetilde{\text{#1}}}\xspace}
\newcommand{\meanErr}{\ensuremath{\Delta f}\xspace}
\newcommand{\better}{\textcolor{goodgreen}{\ding{51}}\xspace}
\newcommand{\worse}{\textcolor{badred}{\ding{55}}\xspace}
\newcommand{\same}{\textcolor{orange}{\ensuremath{\equiv}}\xspace}
\newcommand{\Atwelve}{\^{A}\textsubscript{12}\xspace}
\newcommand{\finalPF}{\ensuremath{\mathit{S}}\xspace}
\newcommand{\fixedPF}{\ensuremath{\widetilde{\finalPF}}\xspace}
\newcommand{\finalSol}{\ensuremath{s}\xspace}
\newcommand{\fixedSol}{\ensuremath{\tilde{s}}\xspace}
\newcommand{\selectedK}{\ensuremath{k}\xspace}
\newcommand{\maxDist}{\ensuremath{\mathtt{MD}}\xspace}

\usepackage[acronyms,shortcuts,section=section,style=super,nopostdot,nogroupskip,nomain,nonumberlist]{glossaries}
\glsdisablehyper

\newacronym{ads}    {ADS}   {automated driving system}
\newacronym{cas}    {CAS}   {collision avoidance system}
\newacronym{cps}    {CPS}   {cyber-physical system}
\newacronym{ea}     {EA}    {evolutionary algorithm}
\newacronym{ed}     {ED}    {Euclidean distance}
\newacronym{ga}     {GA}    {genetic algorithm}
\newacronym{gd}     {GD}    {generational distance}
\newacronym{hv}     {HV}    {hypervolume}
\newacronym{igd}    {IGD}   {inverse generational distance}
\newacronym{iot}    {IoT}   {internet of things}
\newacronym{knn}    {kNN}   {k-nearest neighbours}
\newacronym{moo}    {MOO}   {multi-objective optimisation}
\newacronym{mop}    {MOP}   {multi-objective problem}
\newacronym{mosp}    {MOSP}   {multi-objective search problem}
\newacronym{qi}     {QI}    {quality indicator}
\newacronym{sbt}    {SBT}   {search-based testing}
\newacronym{sed}    {SED}   {standardised Euclidean distance}

\makeatletter
\renewcommand\paragraph{\@startsection{paragraph}{4}{\z@}%
    {-12\p@ \@plus -4\p@ \@minus -4\p@}%
    {-0.5em \@plus -0.22em \@minus -0.1em}%
    {\normalfont\normalsize\bfseries}}
\makeatother

%% file: pythonlistings.tex
\usepackage{listings}

\DeclareFixedFont{\ttb}{T1}{txtt}{bx}{n}{12} 
\DeclareFixedFont{\ttm}{T1}{txtt}{m}{n}{12}  

\usepackage{color}
\definecolor{deepblue}{rgb}{0,0,0.5}
\definecolor{lightblue}{rgb}{0,0,1}
\definecolor{deepred}{rgb}{0.6,0,0}
\definecolor{deepgreen}{rgb}{0,0.5,0}

\definecolor{maroon}{cmyk}{0, 0.87, 0.68, 0.32}
\definecolor{halfgray}{gray}{0.55}
\definecolor{ipython_frame}{RGB}{207, 207, 207}
\definecolor{ipython_bg}{RGB}{255, 255, 255}
\definecolor{ipython_red}{RGB}{186, 33, 33}
\definecolor{ipython_green}{RGB}{0, 128, 0}
\definecolor{ipython_cyan}{RGB}{64, 128, 128}
\definecolor{ipython_purple}{RGB}{170, 34, 255}

\lstdefinelanguage{iPython}{
    morekeywords={access,and,break,class,continue,def,del,elif,else,except,exec,finally,for,from,global,if,import,as,in,is,lambda,not,or,pass,print,raise,return,try,while},%
    morekeywords=[2]{abs,all,any,basestring,bin,bool,bytearray,callable,chr,classmethod,cmp,compile,complex,delattr,dict,dir,divmod,enumerate,eval,execfile,file,filter,float,format,frozenset,getattr,globals,hasattr,hash,help,hex,id,input,int,isinstance,issubclass,iter,len,list,locals,long,map,max,memoryview,min,next,object,oct,open,ord,pow,property,range,raw_input,reduce,reload,repr,reversed,round,set,setattr,slice,sorted,staticmethod,str,sum,super,tuple,type,unichr,unicode,xrange,zip,apply,buffer,coerce,intern,None,Dict,List,Tuple,True,False},%
    sensitive=true,%
    morecomment=[l]\#,%
    morestring=[b]',%
    morestring=[b]",%
    morestring=[s]{'''}{'''},
    morestring=[s]{"""}{"""},
    morestring=[s]{r'}{'},
    morestring=[s]{r"}{"},%
    morestring=[s]{r'''}{'''},%
    morestring=[s]{r"""}{"""},%
    morestring=[s]{u'}{'},
    morestring=[s]{u"}{"},%
    morestring=[s]{u'''}{'''},%
    morestring=[s]{u"""}{"""},%
    literate=
    {á}{{\'a}}1 {é}{{\'e}}1 {í}{{\'i}}1 {ó}{{\'o}}1 {ú}{{\'u}}1
    {Á}{{\'A}}1 {É}{{\'E}}1 {Í}{{\'I}}1 {Ó}{{\'O}}1 {Ú}{{\'U}}1
    {à}{{\`a}}1 {è}{{\`e}}1 {ì}{{\`i}}1 {ò}{{\`o}}1 {ù}{{\`u}}1
    {À}{{\`A}}1 {È}{{\'E}}1 {Ì}{{\`I}}1 {Ò}{{\`O}}1 {Ù}{{\`U}}1
    {ä}{{\"a}}1 {ë}{{\"e}}1 {ï}{{\"i}}1 {ö}{{\"o}}1 {ü}{{\"u}}1
    {Ä}{{\"A}}1 {Ë}{{\"E}}1 {Ï}{{\"I}}1 {Ö}{{\"O}}1 {Ü}{{\"U}}1
    {â}{{\^a}}1 {ê}{{\^e}}1 {î}{{\^i}}1 {ô}{{\^o}}1 {û}{{\^u}}1
    {Â}{{\^A}}1 {Ê}{{\^E}}1 {Î}{{\^I}}1 {Ô}{{\^O}}1 {Û}{{\^U}}1
    {œ}{{\oe}}1 {Œ}{{\OE}}1 {æ}{{\ae}}1 {Æ}{{\AE}}1 {ß}{{\ss}}1
    {ç}{{\c c}}1 {Ç}{{\c C}}1 {ø}{{\o}}1 {å}{{\r a}}1 {Å}{{\r A}}1
    {€}{{\EUR}}1 {£}{{\pounds}}1
    {^}{{{\color{ipython_purple}\^{}}}}1
    {=}{{{\color{ipython_purple}=}}}1
    {+}{{{\color{ipython_purple}+}}}1
    {*}{{{\color{ipython_purple}$^\ast$}}}1
    {/}{{{\color{ipython_purple}/}}}1
    {+=}{{{+=}}}1
    {-=}{{{-=}}}1
    {*=}{{{$^\ast$=}}}1
    {/=}{{{/=}}}1
    {check()}{{{\color{black}{check()}}}}1,
    literate=
     *{-}{{{\color{ipython_purple}-}}}1
     {--}{{{\color{ipython_purple}\textendash}}}2
     {?}{{{\color{ipython_purple}?}}}1
     {\&}{{{\color{ipython_purple}\&}}}1
     {|}{{{\color{ipython_purple}|}}}1,
    identifierstyle=\color{black}\ttfamily,
    commentstyle=\color{ipython_cyan}\ttfamily,
    stringstyle=\color{ipython_red}\ttfamily,
    keepspaces=true,
    showspaces=false,
    showstringspaces=false,
    rulecolor=\color{ipython_frame},
    frame=single,
    frameround={t}{t}{t}{t},
    framexleftmargin=2em,
    xleftmargin=2em,
    numbers=left,
    numberstyle=\tiny\color{halfgray},
    backgroundcolor=\color{ipython_bg},
    basicstyle=\linespread{1.05}\scriptsize,
    keywordstyle=\color{ipython_green}\ttfamily,
}

\newcommand{\py}[2][\small]{\lstinline[language=iPython,basicstyle=#1]{#2}}

%% file: tables/tableStatTestsFiltered.tex
\begin{table}[!tb]
\caption{Comparison between the \ac{knn}-based settings \knnAvg{\selectedK}{\maxDist} and the baseline \baseline.}
\label{table:expResults}
\centering
\begin{subtable}[t]{0.15\textwidth}
\caption{No noise}
\label{tab:resultsNoise0}
\resizebox{\textwidth}{!}{
\begin{tabular}{lccc}
\toprule
App. & \meanQI{\HV} & \meanQI{\IGD} & \meanErr\\
\midrule
\knnAvg{10}{0.25} & \same & \worse & \worse\\
\knnAvg{10}{0.5} & \worse & \worse & \worse\\
\knnAvg{10}{1.0} & \worse & \worse & \worse\\
\knnAvg{10}{2.0} & \worse & \worse & \worse\\
\knnAvg{10}{4.0} & \worse & \worse & \worse\\
\knnAvg{25}{0.25} & \worse & \worse & \worse\\
\knnAvg{25}{0.5} & \worse & \worse & \worse\\
\knnAvg{25}{1.0} & \worse & \worse & \worse\\
\knnAvg{25}{2.0} & \worse & \worse & \worse\\
\knnAvg{25}{4.0} & \worse & \worse & \worse\\
\knnAvg{50}{0.25} & \worse & \worse & \worse\\
\knnAvg{50}{0.5} & \worse & \worse & \worse\\
\knnAvg{50}{1.0} & \worse & \worse & \worse\\
\knnAvg{50}{2.0} & \worse & \worse & \worse\\
\knnAvg{50}{4.0} & \worse & \worse & \worse\\
\knnAvg{100}{0.25} & \worse & \worse & \worse\\
\knnAvg{100}{0.5} & \worse & \worse & \worse\\
\knnAvg{100}{1.0} & \worse & \worse & \worse\\
\knnAvg{100}{2.0} & \worse & \worse & \worse\\
\knnAvg{100}{4.0} & \worse & \worse & \worse\\
\knnAvg{1000}{0.25} & \worse & \worse & \worse\\
\knnAvg{1000}{0.5} & \worse & \worse & \worse\\
\knnAvg{1000}{1.0} & \worse & \worse & \worse\\
\knnAvg{1000}{2.0} & \worse & \worse & \worse\\
\knnAvg{1000}{4.0} & \worse & \worse & \worse\\
\bottomrule
\end{tabular}}
\end{subtable}~
\begin{subtable}[t]{0.15\textwidth}
\caption{\noise{0.05}}
\label{tab:resultsNoise0.05}
\resizebox{\textwidth}{!}{
\begin{tabular}{lccc}
\toprule
App. & \meanQI{\HV} & \meanQI{\IGD} & \meanErr\\
\midrule
\knnAvg{10}{0.25} & \worse & \same & \better\\
\knnAvg{10}{0.5} & \same & \same & \better\\
\knnAvg{10}{1.0} & \worse & \worse & \better\\
\knnAvg{10}{2.0} & \worse & \worse & \better\\
\knnAvg{10}{4.0} & \worse & \worse & \better\\
\knnAvg{25}{0.25} & \same & \same & \better\\
\knnAvg{25}{0.5} & \worse & \worse & \better\\
\knnAvg{25}{1.0} & \worse & \worse & \better\\
\knnAvg{25}{2.0} & \worse & \worse & \better\\
\knnAvg{25}{4.0} & \worse & \worse & \same\\
\knnAvg{50}{0.25} & \same & \same & \better\\
\knnAvg{50}{0.5} & \same & \same & \better\\
\knnAvg{50}{1.0} & \worse & \worse & \better\\
\knnAvg{50}{2.0} & \worse & \worse & \same\\
\knnAvg{50}{4.0} & \worse & \worse & \same\\
\knnAvg{100}{0.25} & \same & \same & \better\\
\knnAvg{100}{0.5} & \worse & \worse & \better\\
\knnAvg{100}{1.0} & \worse & \worse & \same\\
\knnAvg{100}{2.0} & \worse & \worse & \same\\
\knnAvg{100}{4.0} & \worse & \worse & \same\\
\knnAvg{1000}{0.25} & \same & \same & \better\\
\knnAvg{1000}{0.5} & \worse & \worse & \better\\
\knnAvg{1000}{1.0} & \worse & \worse & \same\\
\knnAvg{1000}{2.0} & \worse & \worse & \worse\\
\knnAvg{1000}{4.0} & \worse & \worse & \worse\\
\bottomrule
\end{tabular}}
\end{subtable}~
\begin{subtable}[t]{0.15\textwidth}
\caption{\noise{0.1}}
\label{tab:resultsNoise0.1}
\resizebox{\textwidth}{!}{
\begin{tabular}{lccc}
\toprule
App. & \meanQI{\HV} & \meanQI{\IGD} & \meanErr\\
\midrule
\knnAvg{10}{0.25} & \same & \better & \better\\
\knnAvg{10}{0.5} & \same & \same & \better\\
\knnAvg{10}{1.0} & \same & \same & \better\\
\knnAvg{10}{2.0} & \worse & \worse & \better\\
\knnAvg{10}{4.0} & \worse & \worse & \better\\
\knnAvg{25}{0.25} & \same & \better & \better\\
\knnAvg{25}{0.5} & \same & \same & \better\\
\knnAvg{25}{1.0} & \worse & \same & \better\\
\knnAvg{25}{2.0} & \worse & \worse & \better\\
\knnAvg{25}{4.0} & \worse & \worse & \better\\
\knnAvg{50}{0.25} & \same & \same & \better\\
\knnAvg{50}{0.5} & \same & \same & \better\\
\knnAvg{50}{1.0} & \same & \same & \better\\
\knnAvg{50}{2.0} & \worse & \worse & \better\\
\knnAvg{50}{4.0} & \worse & \worse & \better\\
\knnAvg{100}{0.25} & \same & \same & \better\\
\knnAvg{100}{0.5} & \same & \same & \better\\
\knnAvg{100}{1.0} & \same & \same & \better\\
\knnAvg{100}{2.0} & \worse & \worse & \better\\
\knnAvg{100}{4.0} & \worse & \worse & \better\\
\knnAvg{1000}{0.25} & \same & \better & \better\\
\knnAvg{1000}{0.5} & \same & \same & \better\\
\knnAvg{1000}{1.0} & \same & \same & \better\\
\knnAvg{1000}{2.0} & \worse & \worse & \same\\
\knnAvg{1000}{4.0} & \worse & \worse & \worse\\
\bottomrule
\end{tabular}}
\end{subtable}~
\begin{subtable}[t]{0.15\textwidth}
\caption{\noise{0.25}}
\label{tab:resultsNoise0.25}
\resizebox{\textwidth}{!}{
\begin{tabular}{lccc}
\toprule
App. & \meanQI{\HV} & \meanQI{\IGD} & \meanErr\\
\midrule
\knnAvg{10}{0.25} & \same & \same & \better\\
\knnAvg{10}{0.5} & \same & \better & \better\\
\knnAvg{10}{1.0} & \same & \better & \better\\
\knnAvg{10}{2.0} & \same & \better & \better\\
\knnAvg{10}{4.0} & \worse & \same & \better\\
\knnAvg{25}{0.25} & \same & \same & \better\\
\knnAvg{25}{0.5} & \same & \same & \better\\
\knnAvg{25}{1.0} & \same & \same & \better\\
\knnAvg{25}{2.0} & \worse & \same & \better\\
\knnAvg{25}{4.0} & \worse & \same & \better\\
\knnAvg{50}{0.25} & \better & \better & \better\\
\knnAvg{50}{0.5} & \same & \same & \better\\
\knnAvg{50}{1.0} & \same & \better & \better\\
\knnAvg{50}{2.0} & \worse & \same & \better\\
\knnAvg{50}{4.0} & \worse & \same & \better\\
\knnAvg{100}{0.25} & \same & \better & \better\\
\knnAvg{100}{0.5} & \same & \better & \better\\
\knnAvg{100}{1.0} & \same & \better & \better\\
\knnAvg{100}{2.0} & \same & \same & \better\\
\knnAvg{100}{4.0} & \worse & \same & \better\\
\knnAvg{1000}{0.25} & \same & \better & \better\\
\knnAvg{1000}{0.5} & \same & \better & \better\\
\knnAvg{1000}{1.0} & \same & \better & \better\\
\knnAvg{1000}{2.0} & \same & \better & \better\\
\knnAvg{1000}{4.0} & \same & \same & \better\\
\bottomrule
\end{tabular}}
\end{subtable}~
\begin{subtable}[t]{0.15\textwidth}
\caption{\noise{0.5}}
\label{tab:resultsNoise0.5}
\resizebox{\textwidth}{!}{
\begin{tabular}{lccc}
\toprule
App. & \meanQI{\HV} & \meanQI{\IGD} & \meanErr\\
\midrule
\knnAvg{10}{0.25} & \same & \same & \better\\
\knnAvg{10}{0.5} & \same & \same & \better\\
\knnAvg{10}{1.0} & \same & \same & \better\\
\knnAvg{10}{2.0} & \same & \same & \better\\
\knnAvg{10}{4.0} & \same & \same & \better\\
\knnAvg{25}{0.25} & \same & \same & \better\\
\knnAvg{25}{0.5} & \same & \same & \better\\
\knnAvg{25}{1.0} & \same & \better & \better\\
\knnAvg{25}{2.0} & \same & \same & \better\\
\knnAvg{25}{4.0} & \same & \same & \better\\
\knnAvg{50}{0.25} & \same & \same & \better\\
\knnAvg{50}{0.5} & \same & \same & \better\\
\knnAvg{50}{1.0} & \same & \same & \better\\
\knnAvg{50}{2.0} & \same & \same & \better\\
\knnAvg{50}{4.0} & \same & \same & \better\\
\knnAvg{100}{0.25} & \same & \same & \better\\
\knnAvg{100}{0.5} & \same & \same & \better\\
\knnAvg{100}{1.0} & \same & \same & \better\\
\knnAvg{100}{2.0} & \same & \same & \better\\
\knnAvg{100}{4.0} & \same & \same & \better\\
\knnAvg{1000}{0.25} & \same & \same & \better\\
\knnAvg{1000}{0.5} & \same & \same & \better\\
\knnAvg{1000}{1.0} & \same & \better & \better\\
\knnAvg{1000}{2.0} & \same & \better & \better\\
\knnAvg{1000}{4.0} & \same & \better & \better\\
\bottomrule
\end{tabular}}
\end{subtable}
\begin{subtable}[t]{0.15\textwidth}
\caption{All}
\label{tab:resultsAllNoise}
\resizebox{\textwidth}{!}{
\begin{tabular}{lccc}
\toprule
App. & \meanQI{\HV} & \meanQI{\IGD} & \meanErr\\
\midrule
\knnAvg{10}{0.25} & \same & \better & \better\\
\knnAvg{10}{0.5} & \same & \better & \better\\
\knnAvg{10}{1.0} & \worse & \same & \better\\
\knnAvg{10}{2.0} & \worse & \same & \better\\
\knnAvg{10}{4.0} & \worse & \worse & \better\\
\knnAvg{25}{0.25} & \same & \better & \better\\
\knnAvg{25}{0.5} & \same & \same & \better\\
\knnAvg{25}{1.0} & \worse & \same & \better\\
\knnAvg{25}{2.0} & \worse & \worse & \better\\
\knnAvg{25}{4.0} & \worse & \worse & \better\\
\knnAvg{50}{0.25} & \same & \better & \better\\
\knnAvg{50}{0.5} & \same & \same & \better\\
\knnAvg{50}{1.0} & \worse & \same & \better\\
\knnAvg{50}{2.0} & \worse & \worse & \better\\
\knnAvg{50}{4.0} & \worse & \worse & \better\\
\knnAvg{100}{0.25} & \same & \same & \better\\
\knnAvg{100}{0.5} & \same & \same & \better\\
\knnAvg{100}{1.0} & \worse & \same & \better\\
\knnAvg{100}{2.0} & \worse & \worse & \better\\
\knnAvg{100}{4.0} & \worse & \worse & \better\\
\knnAvg{1000}{0.25} & \same & \better & \better\\
\knnAvg{1000}{0.5} & \same & \same & \better\\
\knnAvg{1000}{1.0} & \same & \same & \better\\
\knnAvg{1000}{2.0} & \worse & \same & \better\\
\knnAvg{1000}{4.0} & \worse & \worse & \same\\
\bottomrule
\end{tabular}}
\end{subtable}
\caption*{\footnotesize Legend: \same: there is no statistically significant difference between \knnAvg{\selectedK}{\maxDist} and \baseline.\\ \better: \knnAvg{\selectedK}{\maxDist} is statistically significantly better. \quad \worse: \baseline is statistically significantly better.}
\end{table}